\documentclass{bmvc2k}

\title{Gradient Weighted Superpixels for Interpretability in CNNs}

\addauthor{Thomas Hartley}{hartleytw@cardiff.ac.uk}{1}
\addauthor{Kirill Sidorov}{sidorovk@cardiff.ac.uk}{1}
\addauthor{Christopher Willis}{chris.willis@baesystems.com}{2}
\addauthor{David Marshall}{marshallad@cardiff.ac.uk}{1}

\addinstitution{
 School of Computer Science\\
 Cardiff University\\
 Cardiff, UK
}
\addinstitution{
 BAE Systems Applied Intelligence\\
 Chelmsford Technology Park,\\
 Great Baddow, UK
}

\runninghead{Hartley, Sidorov, Willis, Marshall}{Gradient Weighted Superpixels}
\pdfoutput=1

\def\etal{\emph{et al}\bmvaOneDot}
\usepackage{multirow}
\usepackage{todonotes}

\begin{document}

\maketitle

\begin{abstract}
As Convolutional Neural Networks embed themselves into our everyday
lives, the need for them to be interpretable increases. However, there is often
a trade-off between methods that are efficient to compute but produce an
explanation that is difficult to interpret, and those that are slow to compute
but provide a more interpretable result. This is particularly challenging in
problem spaces that require a large input volume, especially video which combines
both spatial and temporal dimensions.
In this work we introduce the idea of scoring superpixels through the use of
gradient based pixel scoring techniques. We show qualitatively and quantitatively
that this is able to approximate LIME, in a fraction of the time. We investigate
our techniques using both image
classification, and action recognition networks on large scale
datasets (ImageNet and Kinetics-400 respectively).
\end{abstract}

\section{Introduction}

Convolutional Neural Networks (CNNs) are often described as black boxes due to
the difficulty in explaining how they reach their final output for a given task.
Consequently a number of techniques have been developed to aid in the process of
explainability. These techniques range from the scoring of individual pixels to
reflect their impact on the networks decision making, to the scoring of larger
regions of the image. Scoring larger regions allows for the results to be more
easily interpreted.

A popular technique for explaining images is LIME~\cite{lime}. This uses
superpixels, contiguous regions for visualisation, allowing a level
of interpretability that may not be present in individual pixel scoring.
However, this increased interpretability comes at a cost. The LIME technique
relies on perturbing the input image and repeatedly passing it to the network to
build
an understanding of how important each superpixel region is to the final
classification. This requires multiple perturbed images to be passed through the
network, by default $1000$ in the released code. Having such a computationally
intensive method for ranking regions of an input is problematic when we have a
need for real-time generation of visualisations, or where the input is more
complex
than a 2D image, i.e.\ a 3D spatio-temporal input such as video.

Interpretability of networks is an important area of research, particularly as
techniques come under both increased scrutiny and an expectation they will be
able to give an explanation for their decision (i.e.\ with the recent EU GDPR
coming into effect). Having techniques that are efficient to compute as well as
being easy to interpret are therefore crucial. As interpretability techniques have improved, they have had a tendency to become
more complicated to compute. For example a number of techniques require
multiple passes through the network to explain a single input~
\cite{fong_iccv_2017, lime, zeiler2014visualizing, Sundararajan:2017}. We
propose a method to approximate the results generated by LIME using using a much
less time consuming method.

In this paper we propose a method for weighting superpixels through the use of
aggregated pixel values, achievable in a single forward and backward pass of the
network. We show that our technique is comparable to LIME for a modest number of
passes through the network. We also show how this technique can be extended for
use in spatiotemporal inputs, allowing a novel method of explaining action
recognition networks to be developed.

\section{Related Work}

A number of works have previously attempted to explain how and why a network has
made its decision based on the input space. Initially this work was based on back
propagating the gradients from the output to the input pixels~
\cite{simonyan2013deep, zeiler2014visualizing}. These techniques were built upon
further with the use of guided backpropagation~\cite{springenberg2014striving},
then further expanded by combining the gradient with the
activations during backpropagation in works such as Layer-wise Relevance
Propagation (LRP)~\cite{bach2015pixel}, Deep Taylor~
\cite{montavon2017explaining}, and Excitation Backprop~\cite{zhang2018top}.
Integrated Gradients~\cite{pmlr-v70-shrikumar17a} propose that instead of using
a single input, it is better to have a range of scaled inputs (i.e.\ from zeros
to the original input values) and integrate the corresponding gradients. This
work also introduced gradient $\odot$ input as a visualisation method.
Class Activation Maps~\cite{zhou2015cnnlocalization} (CAMs) allow networks with
a Global Average Pooling (GAP) layer to localise discriminative regions within
the input space. This was generalised in Grad-CAM~\cite{Selvaraju_2017_ICCV} to
allow networks without a GAP layer to produce class activation maps through the
visualisation of the final activation map weighted with the mean gradients. The
technique was adapted with Grad-CAM++~\cite{grad_cam++} which aimed to
improve Grad-CAMs localisation ability. 

Methods have been developed that treat the network as a black box and perturb
the image space to discover how the network makes it decision. An early
example of this was occlusion maps~\cite{zeiler2014visualizing}, which iterates
a blank patch over the input space and stores the softmax score for each
position. As the blank patch moves over important features, the softmax score
should drop allowing a visualisation to be built showing which pixels are
important to the network.
A more recent example of a black box technique is LIME~\cite{lime} which
segments the input space using superpixels before
perturbing them by turning superpixels on or off (i.e.~setting the superpixel
to the median value of the input image). A linear regression model is then
trained on the
perturbations and corresponding scores. By interpreting this local model,
superpixels can be selected that indicate the important regions of the image.
This idea of perturbing the input space was investigated further in the work
by Fong~\etal~\cite{fong_iccv_2017} which again treats the network as a black
box but attempts to learn a mask that maximally suppresses the softmax score for
the given input.





\begin{figure*}[]
  \centering
  \includegraphics[width=\textwidth]{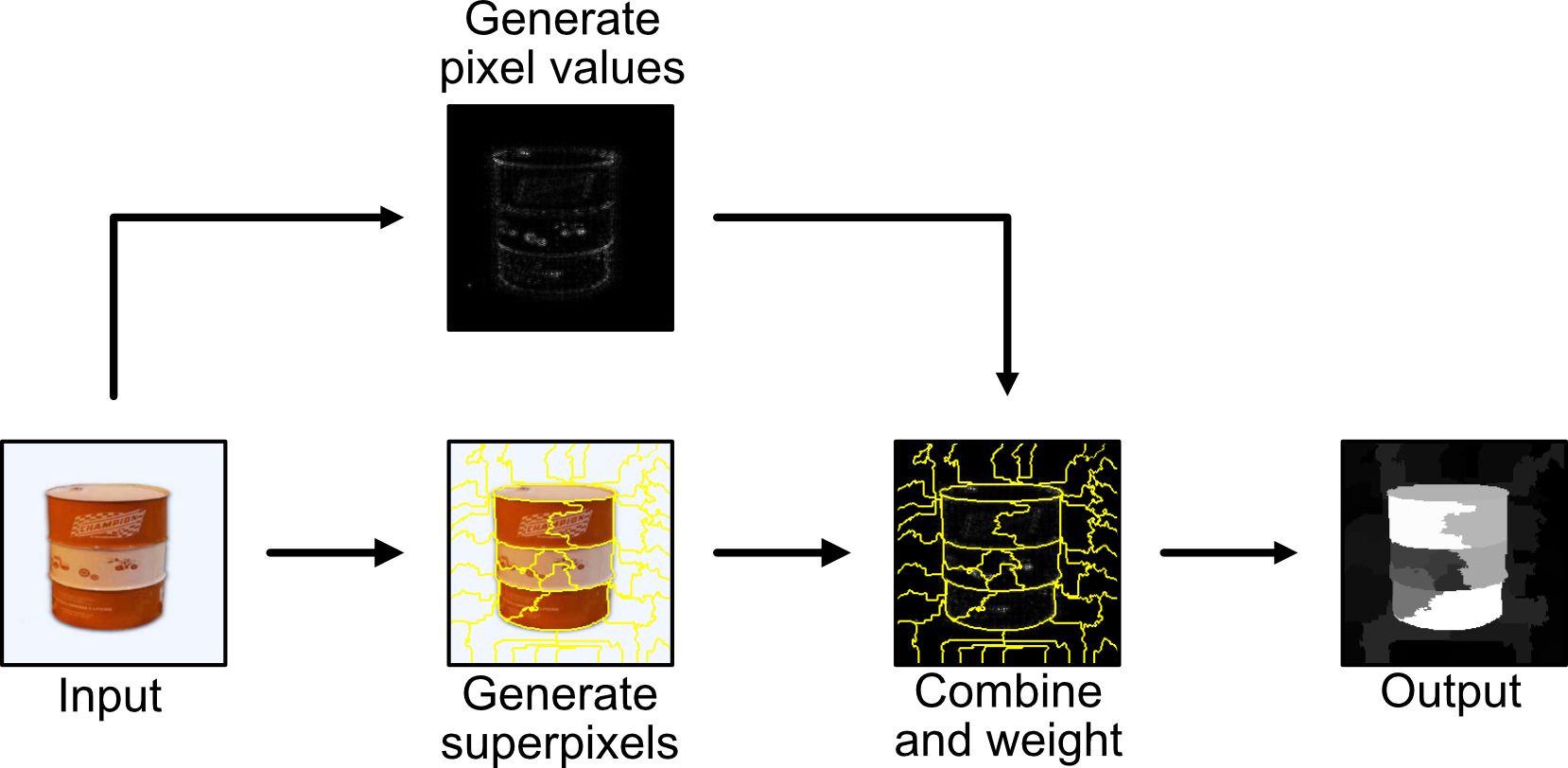}
     \caption{A simple overview of our proposed method.}
  \label{fig:overview}
\end{figure*}
\section{Proposal} Our proposed method is an alternative to the time consuming
method used by LIME. Rather than generating a number of perturbed images and
seeing how the network reacts, we suggest generating the superpixels in the same
way but then weighting each one using the values from a pixel scoring method.
Previous interpretability
techniques have produced saliency maps based on backpropagating gradients back
to the initial input to indicate which pixels are important and which are not.
With a score for each pixel we suggest using the values contained within each
superpixel to provide an overall weight. An overview of the proposed technique
can be found in Figure~\ref{fig:overview}. In this way we are able to approximate
LIME~\cite{lime} with its interpretable superpixels with a reduced computational
footprint. 

\begin{figure*}[]
  \centering
  \begin{tabular}{ccccc}

    \multicolumn{5}{c}{VGG16}\\
    \cline{1-5}

    Input & Vanilla & Guided Vanilla & I$\odot$G & Activation\\
    \raisebox{-0.5\height}{\frame{\includegraphics[width=19mm]
    {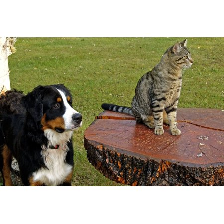}}}&
    \raisebox{-0.5\height}{\includegraphics[width=19mm]
    {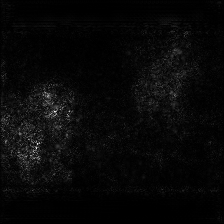}}&
    \raisebox{-0.5\height}{\includegraphics[width=19mm]
    {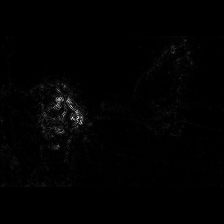}}&
    \raisebox{-0.5\height}{\includegraphics[width=19mm]
    {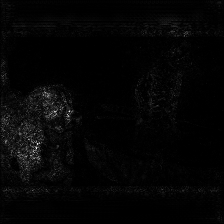}}&
    \raisebox{-0.5\height}{\includegraphics[width=19mm]
    {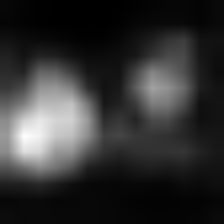}}\\
    Grad-CAM & Guided Grad & Grad-CAM++ & A$\odot$G-CAM & Guided A$\odot$G\\
    \raisebox{-0.5\height}{\includegraphics[width=19mm]
    {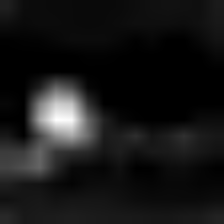}}&
    \raisebox{-0.5\height}{\includegraphics[width=19mm]
    {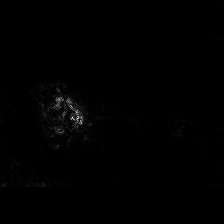}}&
    \raisebox{-0.5\height}{\includegraphics[width=19mm]
    {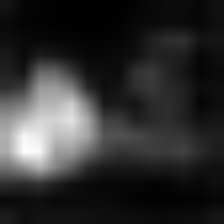}}&
    \raisebox{-0.5\height}{\includegraphics[width=19mm]
    {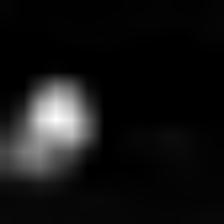}}&
    \raisebox{-0.5\height}{\includegraphics[width=19mm]
    {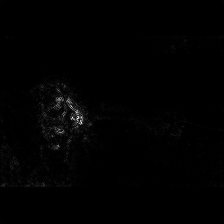}}\\
    &&&&\\


  \end{tabular}
  \caption{Examples of the nine methods investigated using PyTorch's~\cite{paszke2017automatic} pretrained
  VGG16 network. The input
  image is classified as a dog suggesting that all the techniques tried are able
  to correctly locate and explain the dog region in some way.}
  \label{fig:example_methods}
\end{figure*}

We have specifically chosen pixel scoring methods that require at most a single
forward and backward pass through the network. We investigate the following
techniques for producing pixel scores, an example of each are shown in Figure~
\ref{fig:example_methods}:
\begin{description}
\item [Vanilla~\cite{simonyan2013deep}] Pixel scores produced by back
propagating gradients from the softmax layer (for a given class) to the input. 

\item [Guided Vanilla~\cite{springenberg2014striving}] As with vanilla
backpropagation except only positive gradients are backpropagated. 

\item [Input$\odot$Gradient~\cite{pmlr-v70-shrikumar17a}] The product of the
input image and the vanilla backpropagation.

\item [ReLU Activation] The rectified activation map produced by the final
convolution layer. This is resized to match the input dimensions.

\item [Grad-CAM~\cite{Selvaraju_2017_ICCV}] The weighted rectified activation
map produced by the final convolution layer. Here the weights are produced from
the mean of the gradients at the final convolution layer.

\item [Guided Grad-CAM~\cite{Selvaraju_2017_ICCV}] The product of guided
backpropagation and Grad-CAM.

\item [Grad-CAM++~\cite{grad_cam++}] A variant of Grad-CAM that uses a weighted
combination of the positive partial derivatives to inform the weights.

\item [Act$\odot$Grad-CAM] The weighted rectified activation map produced by the final
convolution layer. Here the weights are produced from the mean of the
product of the activations and gradients at the final convolution layer.

\item [Guided Act$\odot$Grad-CAM] The product of guided backpropagation and
Act$\odot$Grad-CAM.

\end{description}

For each method we sum the absolute values of each pixel located within its
respective superpixel to generate that superpixel's weight. Other methods of
distilling the pixel values into superpixel scores were investigated, however
the sum of absolute values was found to be superior.

\subsection{Generating Explanations}

\begin{figure*}[h!]
  \centering
  \begin{tabular}{cccc}
    Input & Superpixels & Dog & Cat\\
    \raisebox{-0.5\height}{\frame{\includegraphics[width=23mm]
    {media/method_examples/input.png}}}&
    \raisebox{-0.5\height}{\frame{\includegraphics[width=23mm]
    {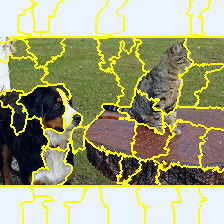}}}&
    \raisebox{-0.5\height}{\includegraphics[width=23mm]
    {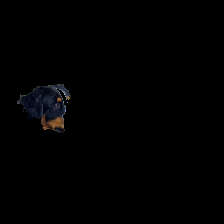}}&
    \raisebox{-0.5\height}{\includegraphics[width=23mm]
    {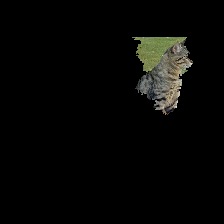}}\\
  \end{tabular}
  \caption{An example of explanations generated for the classes ``Bernese
  mountain dog'' and ``Tabby''. Here we visualise the most important superpixel
  for each class.}
  \label{fig:cat_dog}
\end{figure*}

A benefit of using LIME to explain images is that it allows the superpixels
that contribute to a given class to be highlighted. By visualising the
superpixels that contribute to that class, a greater degree of trust can
be
gained in the network~\cite{lime}. To allow our method to explain different
classes, we must alter the class that we backpropagate from. This method
works for all pixel scoring techniques except for the raw activations. An
example is shown in Figure~\ref{fig:cat_dog}.



\subsection{Extension to Video}


An interesting extension to this technique is in its use for networks that
require temporal inputs. Typically these networks take longer to process a
block of temporal information compared to a single image, this amplifies the
time issues faced by LIME. Temporal networks are often difficult to visualise
using techniques developed for image networks. For example, with techniques that
derive their understanding of the network from the final convolution layer 
(i.e.\ CAM
or Grad-CAM), not only do they have to be resized spatially as with an image,
but also temporally. For example, in the C3D network~\cite{Tran_2015_ICCV} the
initial $16$ frames are compressed to $2$ by the final convolution layer.

\begin{figure}[]
\centering
\footnotesize
\begin{tabular}{rl@{}l@{}l@{}l@{}l@{}l@{}l@{}l@{}l@{}l@{}l@{}l@{}l@{}l@{}l@{}l@{}}
 &
\includegraphics[width=1.5cm]{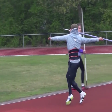}&
\includegraphics[width=1.5cm]{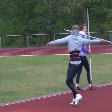}&
\includegraphics[width=1.5cm]{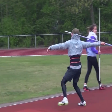}&
\includegraphics[width=1.5cm]{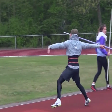}&
\includegraphics[width=1.5cm]{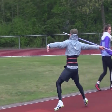}&
\includegraphics[width=1.5cm]{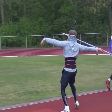}&
\includegraphics[width=1.5cm]{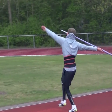}&
\includegraphics[width=1.5cm]{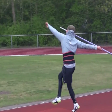}&
\\
 &
\includegraphics[width=1.5cm]{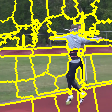}&
\includegraphics[width=1.5cm]{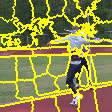}&
\includegraphics[width=1.5cm]{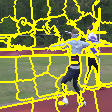}&
\includegraphics[width=1.5cm]{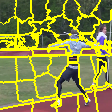}&
\includegraphics[width=1.5cm]{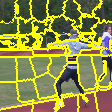}&
\includegraphics[width=1.5cm]{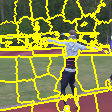}&
\includegraphics[width=1.5cm]{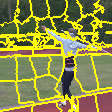}&
\includegraphics[width=1.5cm]{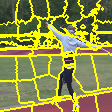}&
\\
 &
\includegraphics[width=1.5cm]{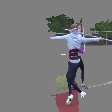}&
\includegraphics[width=1.5cm]{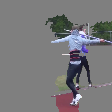}&
\includegraphics[width=1.5cm]{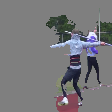}&
\includegraphics[width=1.5cm]{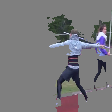}&
\includegraphics[width=1.5cm]{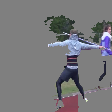}&
\includegraphics[width=1.5cm]{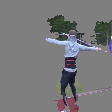}&
\includegraphics[width=1.5cm]{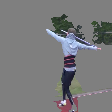}&
\includegraphics[width=1.5cm]{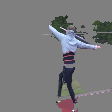}&
\\
 &
\includegraphics[width=1.5cm]{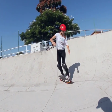}&
\includegraphics[width=1.5cm]{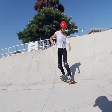}&
\includegraphics[width=1.5cm]{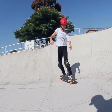}&
\includegraphics[width=1.5cm]{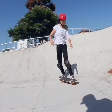}&
\includegraphics[width=1.5cm]{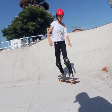}&
\includegraphics[width=1.5cm]{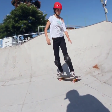}&
\includegraphics[width=1.5cm]{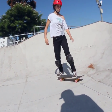}&
\includegraphics[width=1.5cm]{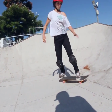}&
\\
 &
\includegraphics[width=1.5cm]{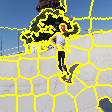}&
\includegraphics[width=1.5cm]{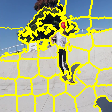}&
\includegraphics[width=1.5cm]{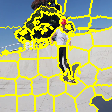}&
\includegraphics[width=1.5cm]{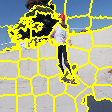}&
\includegraphics[width=1.5cm]{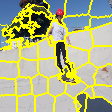}&
\includegraphics[width=1.5cm]{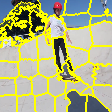}&
\includegraphics[width=1.5cm]{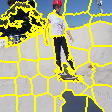}&
\includegraphics[width=1.5cm]{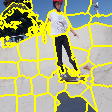}&
\\
 &
\includegraphics[width=1.5cm]{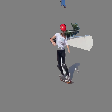}&
\includegraphics[width=1.5cm]{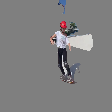}&
\includegraphics[width=1.5cm]{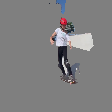}&
\includegraphics[width=1.5cm]{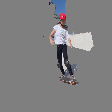}&
\includegraphics[width=1.5cm]{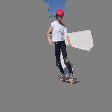}&
\includegraphics[width=1.5cm]{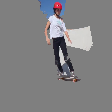}&
\includegraphics[width=1.5cm]{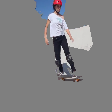}&
\includegraphics[width=1.5cm]{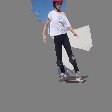}&
\\

\end{tabular}
\caption{Example of technique applied to videos for the classes ``Javelin
Throw'' and ``Skateboarding'' respectively.
Showing alternating frames from a 16 frame input, the generated superpixels and
the top 5 superpixels.}
\label{figure:video_example}

\end{figure}

We propose using a segmentation technique that accommodates 3D volumes to allow
our segments to extend through time. Using the pixel scores from the previously
discussed techniques, we weight our 3D segments in the same manner as images. An
example of our technique is shown in Figure~\ref{figure:video_example}.

\section{Experiments}

We baseline our proposed technique against both LIME and the random ranking of
superpixels. As we are aiming to find an efficient alternative to LIME we
perform a number of baseline experiments with
a differing number of sample inputs, i.e.\ the number of perturbed images fed
into the network for each image. This allows us to understand if and where a
trade off point between using our proposed technique and LIME exists.

All work is implemented in PyTorch~\cite{paszke2017automatic}. For image
classification models we use the
pretrained VGG16~\cite{Simonyan15} and ResNet50~\cite{He_2016_CVPR} networks
with ImageNet. For action
recognition we use our own implementation of C3D loaded with weights for the
Kinetics-400
dataset~\cite{kay2017kinetics, carreira2017quo} released by the authors
alongside their R(2+1)D model~\cite{r2plus1d_cvpr18}. Whilst C3D is
not state of the art it still uses a similar input size (i.e.
112$\times$112$\times$16) as more
more recent models. Generation of superpixels
is achieved using QuickShift~\cite{vedaldi2008quick} for image classification task and SLIC~
\cite{achanta2012slic} for action
recognition tasks. We seed the method for generating superpixels to ensure the
generated superpixels are identical across all experiments.

\begin{figure*}[]
  \centering
  \begin{tabular}{c@{}c@{}c@{}c@{}c@{}c@{}c@{}c@{}c@{}c@{}}


    Input & Superpix & Guided & G-CAM & AG-CAM & LIME$_{50}$ & LIME$_{75}$
    &
    LIME$_{100}$& LIME$_{5000}$\\
    \raisebox{-0.5\height}{\frame{\includegraphics[width=14mm]
    {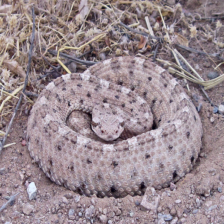}}}&
    \raisebox{-0.5\height}{\frame{\includegraphics[width=14mm]
    {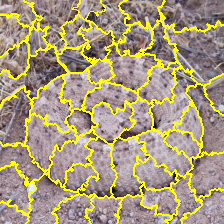}}}&
    \raisebox{-0.5\height}{\frame{\includegraphics[width=14mm]
    {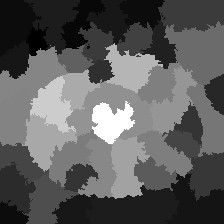}}}&
    \raisebox{-0.5\height}{\frame{\includegraphics[width=14mm]
    {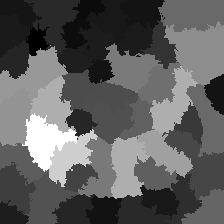}}}&
    \raisebox{-0.5\height}{\frame{\includegraphics[width=14mm]
    {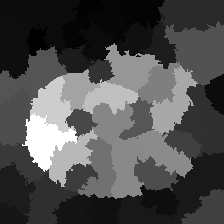}}}&
    \raisebox{-0.5\height}{\frame{\includegraphics[width=14mm]
    {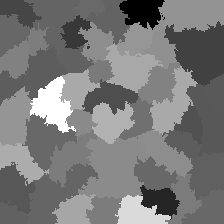}}}&
    \raisebox{-0.5\height}{\frame{\includegraphics[width=14mm]
    {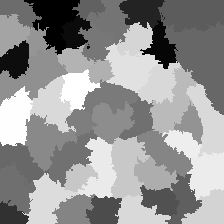}}}&
    \raisebox{-0.5\height}{\frame{\includegraphics[width=14mm]
    {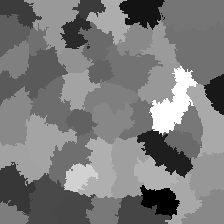}}}&
    \raisebox{-0.5\height}{\frame{\includegraphics[width=14mm]
    {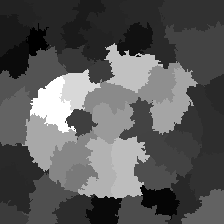}}}\\

    \raisebox{-0.5\height}{\frame{\includegraphics[width=14mm]
    {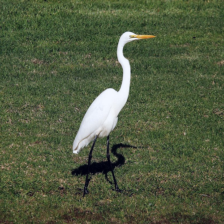}}}&
    \raisebox{-0.5\height}{\frame{\includegraphics[width=14mm]
    {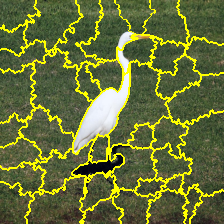}}}&
    \raisebox{-0.5\height}{\frame{\includegraphics[width=14mm]
    {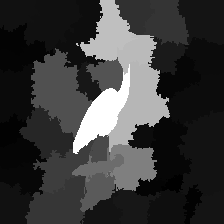}}}&
    \raisebox{-0.5\height}{\frame{\includegraphics[width=14mm]
    {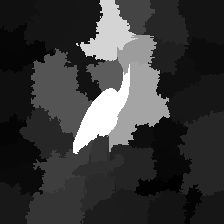}}}&
    \raisebox{-0.5\height}{\frame{\includegraphics[width=14mm]
    {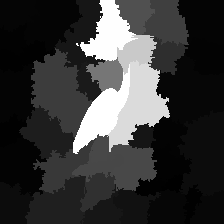}}}&
    \raisebox{-0.5\height}{\frame{\includegraphics[width=14mm]
    {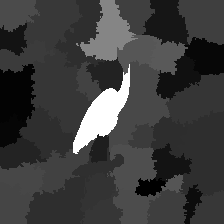}}}&
    \raisebox{-0.5\height}{\frame{\includegraphics[width=14mm]
    {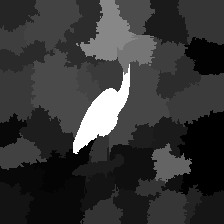}}}&
    \raisebox{-0.5\height}{\frame{\includegraphics[width=14mm]
    {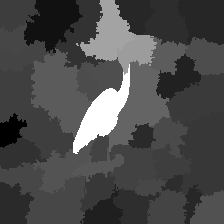}}}&
    \raisebox{-0.5\height}{\frame{\includegraphics[width=14mm]
    {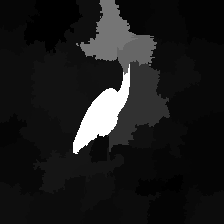}}}\\


    \raisebox{-0.5\height}{\frame{\includegraphics[width=14mm]
    {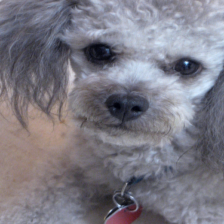}}}&
    \raisebox{-0.5\height}{\frame{\includegraphics[width=14mm]
    {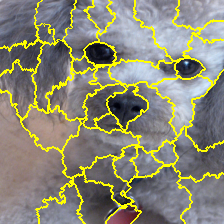}}}&
    \raisebox{-0.5\height}{\frame{\includegraphics[width=14mm]
    {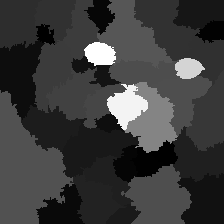}}}&
    \raisebox{-0.5\height}{\frame{\includegraphics[width=14mm]
    {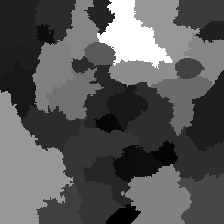}}}&
    \raisebox{-0.5\height}{\frame{\includegraphics[width=14mm]
    {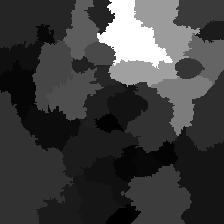}}}&
    \raisebox{-0.5\height}{\frame{\includegraphics[width=14mm]
    {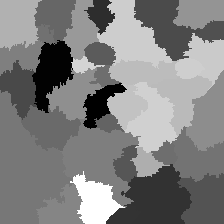}}}&
    \raisebox{-0.5\height}{\frame{\includegraphics[width=14mm]
    {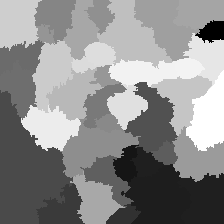}}}&
    \raisebox{-0.5\height}{\frame{\includegraphics[width=14mm]
    {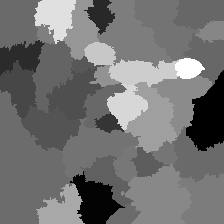}}}&
    \raisebox{-0.5\height}{\frame{\includegraphics[width=14mm]
    {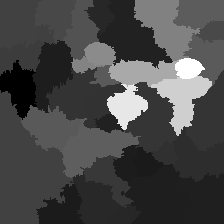}}}\\

    \raisebox{-0.5\height}{\frame{\includegraphics[width=14mm]
    {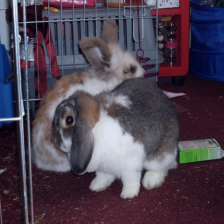}}}&
    \raisebox{-0.5\height}{\frame{\includegraphics[width=14mm]
    {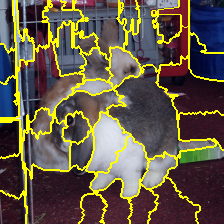}}}&
    \raisebox{-0.5\height}{\frame{\includegraphics[width=14mm]
    {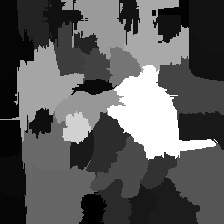}}}&
    \raisebox{-0.5\height}{\frame{\includegraphics[width=14mm]
    {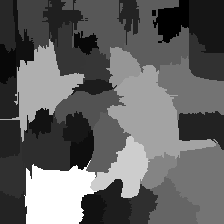}}}&
    \raisebox{-0.5\height}{\frame{\includegraphics[width=14mm]
    {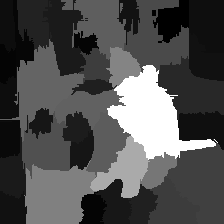}}}&
    \raisebox{-0.5\height}{\frame{\includegraphics[width=14mm]
    {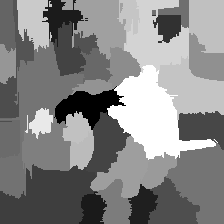}}}&
    \raisebox{-0.5\height}{\frame{\includegraphics[width=14mm]
    {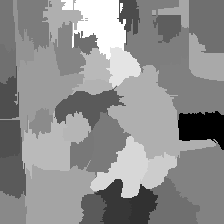}}}&
    \raisebox{-0.5\height}{\frame{\includegraphics[width=14mm]
    {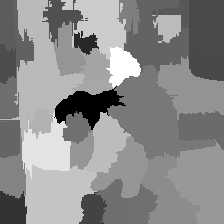}}}&
    \raisebox{-0.5\height}{\frame{\includegraphics[width=14mm]
    {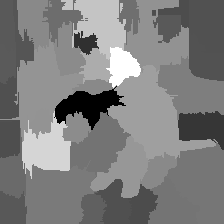}}}\\

    \raisebox{-0.5\height}{\frame{\includegraphics[width=14mm]
    {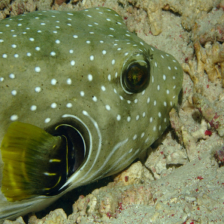}}}&
    \raisebox{-0.5\height}{\frame{\includegraphics[width=14mm]
    {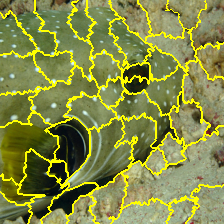}}}&
    \raisebox{-0.5\height}{\frame{\includegraphics[width=14mm]
    {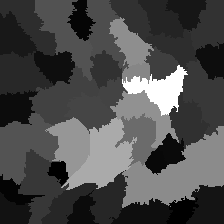}}}&
    \raisebox{-0.5\height}{\frame{\includegraphics[width=14mm]
    {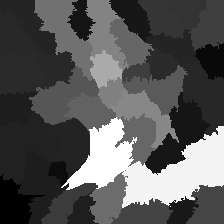}}}&
    \raisebox{-0.5\height}{\frame{\includegraphics[width=14mm]
    {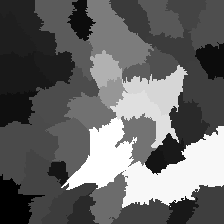}}}&
    \raisebox{-0.5\height}{\frame{\includegraphics[width=14mm]
    {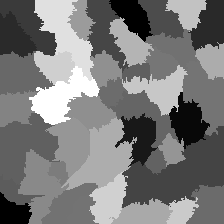}}}&
    \raisebox{-0.5\height}{\frame{\includegraphics[width=14mm]
    {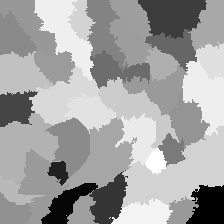}}}&
    \raisebox{-0.5\height}{\frame{\includegraphics[width=14mm]
    {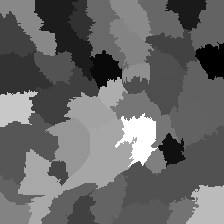}}}&
    \raisebox{-0.5\height}{\frame{\includegraphics[width=14mm]
    {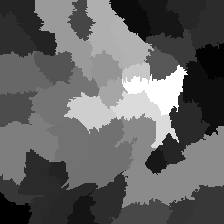}}}\\

    \raisebox{-0.5\height}{\frame{\includegraphics[width=14mm]
    {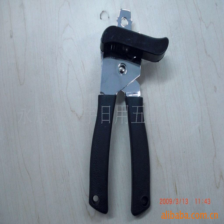}}}&
    \raisebox{-0.5\height}{\frame{\includegraphics[width=14mm]
    {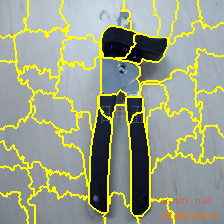}}}&
    \raisebox{-0.5\height}{\frame{\includegraphics[width=14mm]
    {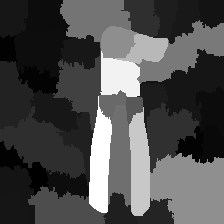}}}&
    \raisebox{-0.5\height}{\frame{\includegraphics[width=14mm]
    {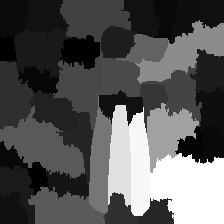}}}&
    \raisebox{-0.5\height}{\frame{\includegraphics[width=14mm]
    {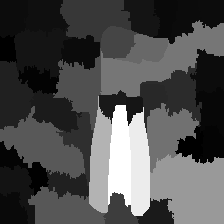}}}&
    \raisebox{-0.5\height}{\frame{\includegraphics[width=14mm]
    {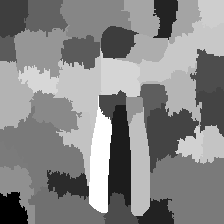}}}&
    \raisebox{-0.5\height}{\frame{\includegraphics[width=14mm]
    {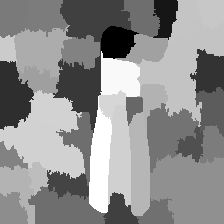}}}&
    \raisebox{-0.5\height}{\frame{\includegraphics[width=14mm]
    {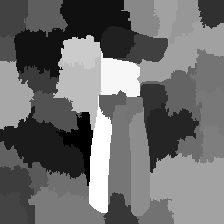}}}&
    \raisebox{-0.5\height}{\frame{\includegraphics[width=14mm]
    {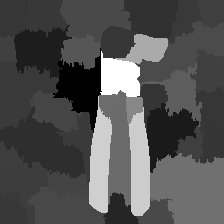}}}\\

    \raisebox{-0.5\height}{\frame{\includegraphics[width=14mm]
    {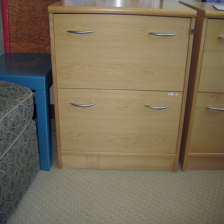}}}&
    \raisebox{-0.5\height}{\frame{\includegraphics[width=14mm]
    {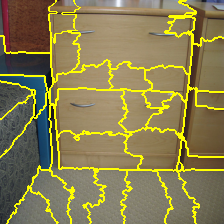}}}&
    \raisebox{-0.5\height}{\frame{\includegraphics[width=14mm]
    {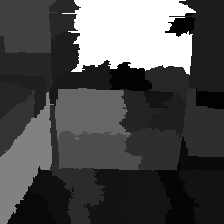}}}&
    \raisebox{-0.5\height}{\frame{\includegraphics[width=14mm]
    {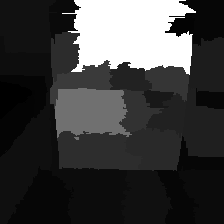}}}&
    \raisebox{-0.5\height}{\frame{\includegraphics[width=14mm]
    {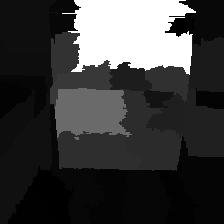}}}&
    \raisebox{-0.5\height}{\frame{\includegraphics[width=14mm]
    {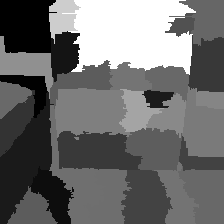}}}&
    \raisebox{-0.5\height}{\frame{\includegraphics[width=14mm]
    {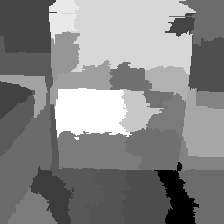}}}&
    \raisebox{-0.5\height}{\frame{\includegraphics[width=14mm]
    {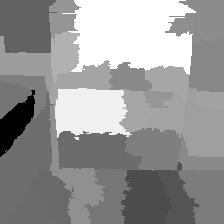}}}&
    \raisebox{-0.5\height}{\frame{\includegraphics[width=14mm]
    {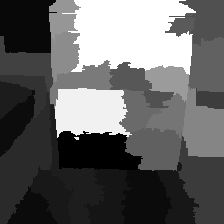}}}\\

    \raisebox{-0.5\height}{\frame{\includegraphics[width=14mm]
    {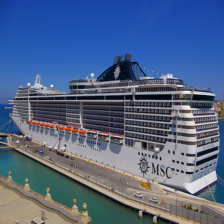}}}&
    \raisebox{-0.5\height}{\frame{\includegraphics[width=14mm]
    {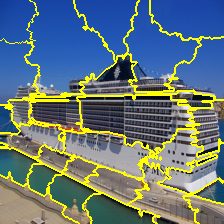}}}&
    \raisebox{-0.5\height}{\frame{\includegraphics[width=14mm]
    {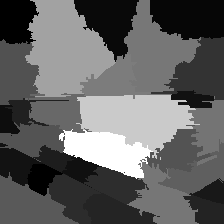}}}&
    \raisebox{-0.5\height}{\frame{\includegraphics[width=14mm]
    {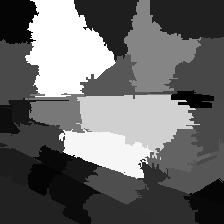}}}&
    \raisebox{-0.5\height}{\frame{\includegraphics[width=14mm]
    {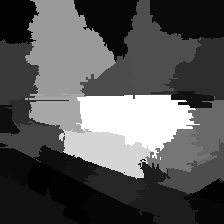}}}&
    \raisebox{-0.5\height}{\frame{\includegraphics[width=14mm]
    {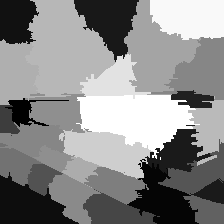}}}&
    \raisebox{-0.5\height}{\frame{\includegraphics[width=14mm]
    {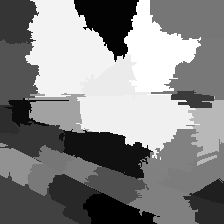}}}&
    \raisebox{-0.5\height}{\frame{\includegraphics[width=14mm]
    {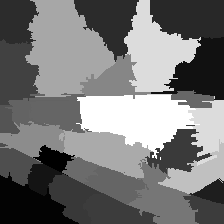}}}&
    \raisebox{-0.5\height}{\frame{\includegraphics[width=14mm]
    {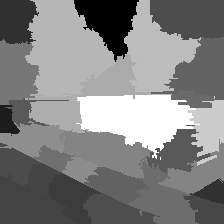}}}\\

    \raisebox{-0.5\height}{\frame{\includegraphics[width=14mm]
    {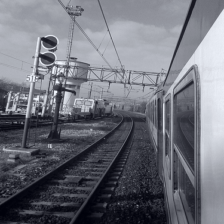}}}&
    \raisebox{-0.5\height}{\frame{\includegraphics[width=14mm]
    {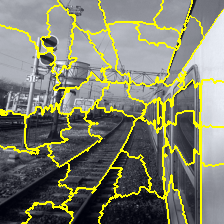}}}&
    \raisebox{-0.5\height}{\frame{\includegraphics[width=14mm]
    {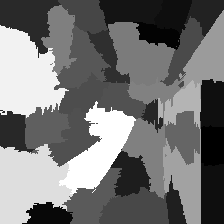}}}&
    \raisebox{-0.5\height}{\frame{\includegraphics[width=14mm]
    {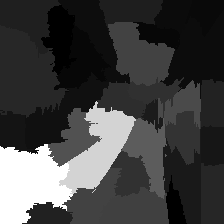}}}&
    \raisebox{-0.5\height}{\frame{\includegraphics[width=14mm]
    {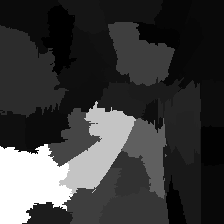}}}&
    \raisebox{-0.5\height}{\frame{\includegraphics[width=14mm]
    {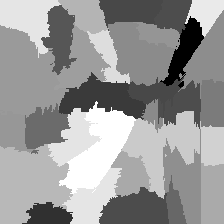}}}&
    \raisebox{-0.5\height}{\frame{\includegraphics[width=14mm]
    {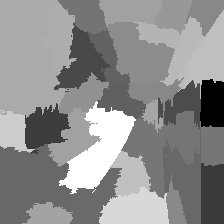}}}&
    \raisebox{-0.5\height}{\frame{\includegraphics[width=14mm]
    {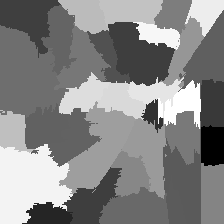}}}&
    \raisebox{-0.5\height}{\frame{\includegraphics[width=14mm]
    {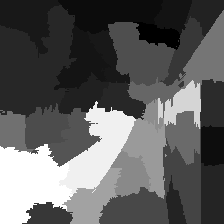}}}\\

        \raisebox{-0.5\height}{\frame{\includegraphics[width=14mm]
    {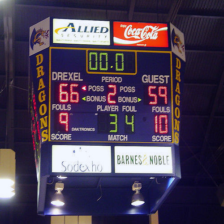}}}&
    \raisebox{-0.5\height}{\frame{\includegraphics[width=14mm]
    {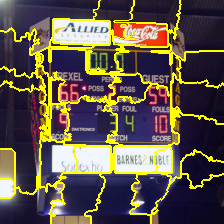}}}&
    \raisebox{-0.5\height}{\frame{\includegraphics[width=14mm]
    {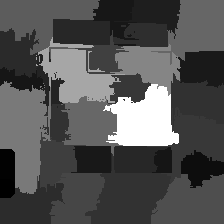}}}&
    \raisebox{-0.5\height}{\frame{\includegraphics[width=14mm]
    {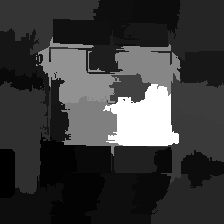}}}&
    \raisebox{-0.5\height}{\frame{\includegraphics[width=14mm]
    {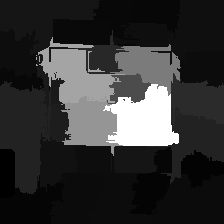}}}&
    \raisebox{-0.5\height}{\frame{\includegraphics[width=14mm]
    {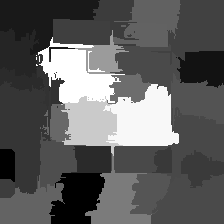}}}&
    \raisebox{-0.5\height}{\frame{\includegraphics[width=14mm]
    {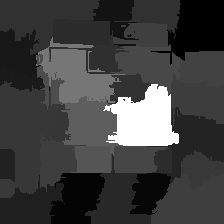}}}&
    \raisebox{-0.5\height}{\frame{\includegraphics[width=14mm]
    {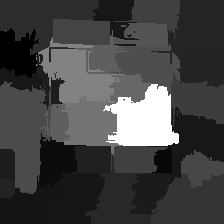}}}&
    \raisebox{-0.5\height}{\frame{\includegraphics[width=14mm]
    {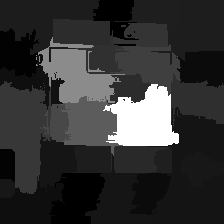}}}\\

    \raisebox{-0.5\height}{\frame{\includegraphics[width=14mm]
    {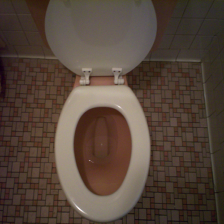}}}&
    \raisebox{-0.5\height}{\frame{\includegraphics[width=14mm]
    {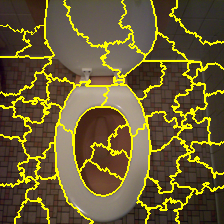}}}&
    \raisebox{-0.5\height}{\frame{\includegraphics[width=14mm]
    {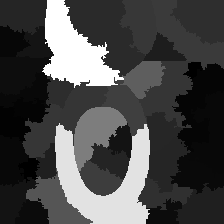}}}&
    \raisebox{-0.5\height}{\frame{\includegraphics[width=14mm]
    {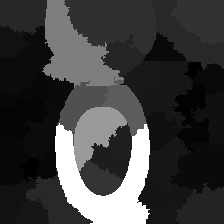}}}&
    \raisebox{-0.5\height}{\frame{\includegraphics[width=14mm]
    {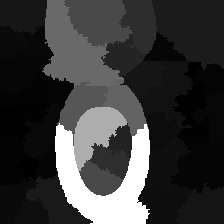}}}&
    \raisebox{-0.5\height}{\frame{\includegraphics[width=14mm]
    {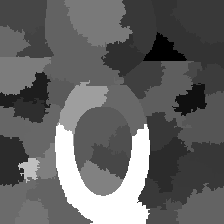}}}&
    \raisebox{-0.5\height}{\frame{\includegraphics[width=14mm]
    {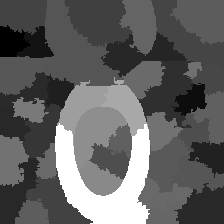}}}&
    \raisebox{-0.5\height}{\frame{\includegraphics[width=14mm]
    {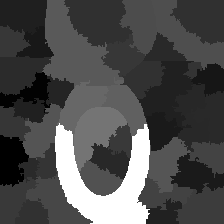}}}&
    \raisebox{-0.5\height}{\frame{\includegraphics[width=14mm]
    {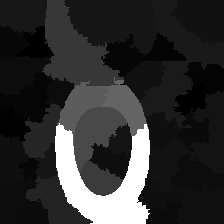}}}\\

    \raisebox{-0.5\height}{\frame{\includegraphics[width=14mm]
    {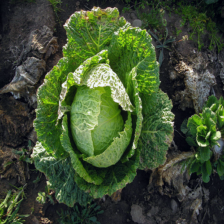}}}&
    \raisebox{-0.5\height}{\frame{\includegraphics[width=14mm]
    {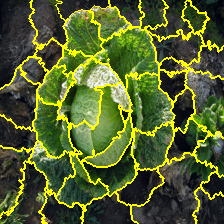}}}&
    \raisebox{-0.5\height}{\frame{\includegraphics[width=14mm]
    {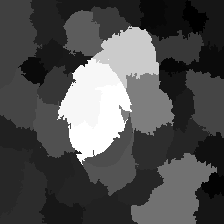}}}&
    \raisebox{-0.5\height}{\frame{\includegraphics[width=14mm]
    {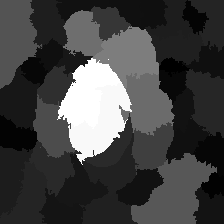}}}&
    \raisebox{-0.5\height}{\frame{\includegraphics[width=14mm]
    {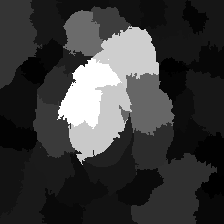}}}&
    \raisebox{-0.5\height}{\frame{\includegraphics[width=14mm]
    {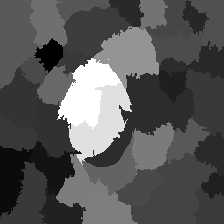}}}&
    \raisebox{-0.5\height}{\frame{\includegraphics[width=14mm]
    {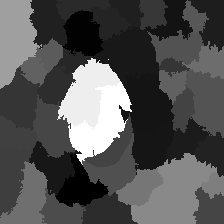}}}&
    \raisebox{-0.5\height}{\frame{\includegraphics[width=14mm]
    {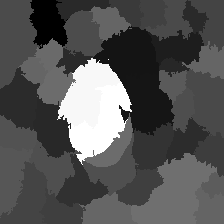}}}&
    \raisebox{-0.5\height}{\frame{\includegraphics[width=14mm]
    {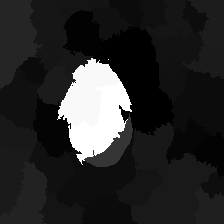}}}\\

  \end{tabular}
  \caption{Examples of three of the better performing weighting methods
  compared to LIME with 50, 75, 100, and 5000 samples. Here we use 5000 samples
  to show how LIME with a much larger number of samples than would normally be
  practical. Here, guided, G-CAM,
   and AG-CAM refer to guided-backpropagation, Grad-CAM and Act$\odot$Grad-CAM
   respectively. These results are generated using VGG16 with ImageNet.}
  \label{fig:lime_comp}
\end{figure*}
\subsection{Superpixel Removal}
We propose experiments to measure how well a technique ranks superpixels. To
understand how accurately the superpixels are ranked we suggest iteratively
removing the highest ranked superpixels from the image and seeing how many can
be removed before the network no longer classifies the image correctly. The
better any technique is at ranking the superpixels the fewer we should be able
to remove before the image is misclassified. Alternatively, we also propose
iteratively removing the lowest ranked superpixels until the image is
misclassified. The more superpixels that can be removed shows how accurate a
technique is at identifying areas of low importance. These results are reported
as the average percentage of superpixels removed over the ImageNet validation
set. An
example of three of the better performing weighting methods compared to LIME is
shown in Figure~\ref{fig:lime_comp}.
Results for this experiment are found in Table~\ref{table:im_results}.
\begin{table}[h]
\centering
\footnotesize
\begin{tabular}{rlcccccccc}
&  & \multicolumn{2}{c}{VGG16} & & \multicolumn{2}{c}{ResNet50} & & 
\multicolumn{2}{c}{C3D}\\
Method  &  & Worst & Best & & Worst & Best & & Worst & Best \\
\cline{1-1} \cline{3-4} \cline{6-7} \cline{9-10} 
Random    &  &  34.48\% &34.50\% &  &40.45\%& 40.47\% &  &  25.49\% &25.66\%\\
LIME 50   &  & 60.78\% & 13.07\% &  & 68.83\% & 15.77\% &  & 42.44\% & 12.14\%\\
LIME 75   &  & 64.02\% & 11.16\% &  & 71.69\% & 13.54\%&  & 45.25\% & 10.30\% \\
LIME 100  &  &  66.15\% &  9.87\% &  & 73.59\% &  12.35\% &  & 47.93\% & 9.12\%\\
LIME 500  &  &  74.79\% &  7.20\% &  &   79.84\% &  9.20\% &&   59.97\% & 
5.92\%\\
LIME 1000 &  &  76.16\% &  6.94\% &  &   80.85\% &  8.81\% &  &  63.20\% &  5.54\%\\
&&&&&&&&&\\
Vanilla           && 61.50\% & 14.21\% && 66.18\%  & 19.36\% &&  45.91\% & 14.95\%\\
Guided Vanilla    && 66.60\% & 11.40\% && 72.01\%  & 14.85\% &&  49.96\% &  \textbf{11.92}\%\\
Input$\odot$Gradient  && 59.38\% & 16.53\% && 64.14\%  & 21.57\% &&  43.90\% & 16.29\%\\
 Grad-CAM         && 65.83\% & 12.24\% &  & 71.84\%  & 14.75\% &  &  48.49\% &  14.69\%\\
Guided Grad-CAM   && 66.31\% & 13.09\% &  & 74.35\%  & \textbf{14.19\%}& & 51.19\% & 12.53\%\\
Grad-CAM++		  && 64.89\% & 12.25\% &  & 69.91\%  & 15.39\% &  &  47.95\% &
14.75\%\\
Act$\odot$Grad-CAM        && 67.28\% & \textbf{11.31\%} &  & 70.54\%  & 15.09\%&  & 
47.85\% &  15.50\%\\
Guided Act$\odot$Grad-CAM && 67.30\% & 11.68\% &  & 73.51\%  & 14.33\% &&  50.99\% & 
13.07\%\\
ReLU Activation   && 63.70\% & 12.64\% &  & 66.26\%  & 17.70\% &  &  42.32\% &
 20.86\%\\ 
\end{tabular}
\caption{Results for superpixel removal where ``Best'' refers to the removal of
the highest ranked superpixels first and ``Worst'' is the removal of the lowest
ranked superpixels first. For ``Best'', a low score is preferable, and ``Worst''
a high score is preferable.}\label{table:im_results}
\end{table}

 These results suggest that all the implemented gradient techniques offer some
 ability to correctly weight the superpixels in a similar way to LIME. All
 techniques investigated, beat the random ranking of superpixels. A number of
 techniques in particular perform well, notably Act$\odot$Grad-CAM for
 VGG-16 and
 Guided Grad-CAM for ResNet50. Conversely with the C3D network we find that
 guided backprop performs the best. This is presumably due to the previously
 discussed problems with expanding CAMS back to the original temporal
 dimension. Overall the most consistent method for all networks is guided
 backpropagation which is only $-0.09\%$ from the best performing VGG16 method,
 and $-0.66\%$ from  ResNet50's best performing method. Interestingly, our
 proposed technique is consistently better at ranking the worst performing
 superpixels. For each model we are able to find a weighting method that beats
 LIME with $100$ samples.


\subsection{Top $k$ Comparison}
Besides comparing how well our proposed technique compares at overall ranking of
superpixels, we also propose an experiment to compare how well we can identify
the most important superpixel. We use LIME with 1000 samples as our
baseline. This is important as, whilst the previous experiment shows how well
our proposed method does in general, the robustness of networks to pixel removal
may distort these results when taken as a whole. In contrast this experiment
allows us to ascertain the proposed methods precision at the superpixel level.
For a given unseen input, if the superpixel ranked as the most important is the
same as
that identified in the top $k$ ranked superpixels using the baseline technique,
we call it correct. Performing
this for all inputs in the validation set gives us an accuracy for each method.
We also compare against LIME with smaller samples. Results
are shown in Table~\ref{table:topk_results}.
\begin{table}[h]
\centering
\footnotesize
\begin{tabular}{rlccccccc}
&  & \multicolumn{3}{c}{$k = 1$} &  & \multicolumn{3}{c}{$k = 5$}\\
Method  &  & VGG16 & ResNet50 & C3D &  & VGG16 & ResNet50 & C3D\\
\cline{1-1} \cline{3-5} \cline{7-9}
LIME 50   &  & 43.71\% & 49.27\% & 23.93\% &  & 69.98\% & 76.44\% & 47.63\% \\
LIME 75   &  & 51.07\% & 57.34\% & 30.40\% &  & 77.18\% & 83.61\% & 55.92\% \\
LIME 100  &  & 56.27\% & 62.59\% & 35.79\% &  & 81.53\% & 87.97\% & 61.53\% \\
LIME 500  &  & 76.82\% & 80.88\% & 61.85\% &  & 95.23\% & 97.62\% & 87.33\% \\
&&&\\
Vanilla           &  & 32.27\% & 27.96\%  & 24.04\% && 65.30\% & 60.34\% &
52.47\%\\ 
Guided Vanilla    &  & \textbf{44.95}\% & 44.48\% & \textbf{33.24}\%  && 78.58\% & 78.09\% &
\textbf{64.05}\% \\
Input$\odot$Gradient  &  & 26.17\% & 23.16\% & 18.75\%  &  & 57.09\% & 53.42\% &
45.35\% \\
Grad-CAM          &  & 41.29\% & 36.75\% & 22.88\%  && 78.36\% & 74.62\% &
52.42\% \\
Guided Grad-CAM   &  & 40.36\% & \textbf{48.71}\% & 31.97\%  && 77.12\% & \textbf{82.93}\% &
63.55\% \\
Grad-CAM++        &  & 39.43\% & 34.37\% & 20.47\%  &  & 75.27\% & 71.16\% &
45.37\% \\
Act$\odot$Grad-CAM        &  & 43.55\% & 34.84\% & 22.31\%  && \textbf{80.09}\% & 72.22\% &
51.41\% \\
Guided Act$\odot$Grad-CAM &  & 43.32\% & 47.35\% & 31.33\%  && 78.99\% & 81.35\% &
63.06\% \\
ReLU Activation   &  & 37.15\% & 27.90\% & 5.53\%  && 72.09\% & 61.62\% &
17.94\% \\

\end{tabular}
\caption{Top k results compared to LIME 1000}\label{table:topk_results}
\end{table}

In these results we can see that again our proposed technique of weighting the
segments is comparable to LIME for around 50 to 75 samples. When trying to find
the most import superpixel chosen by LIME with $1000$ samples within the top $5$
predictions
both VGG16 and C3D results are closer to LIME with $100$ samples. 

\subsection{Time Comparison}

\begin{figure}[h!]
\centering
\includegraphics[width=4cm]{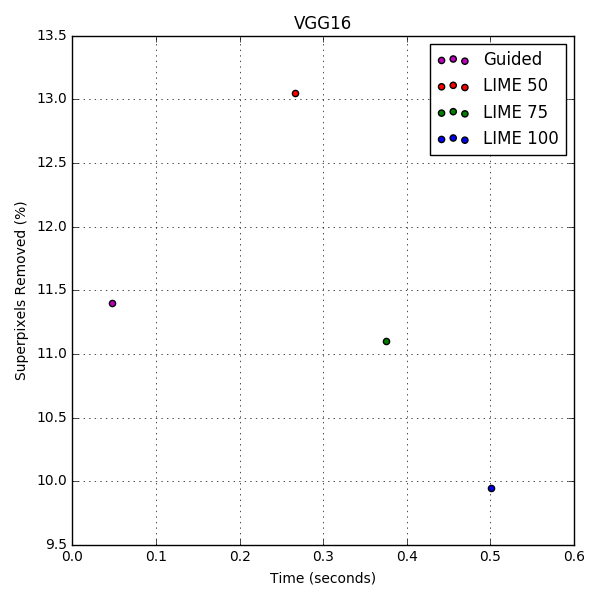}
\includegraphics[width=4cm]{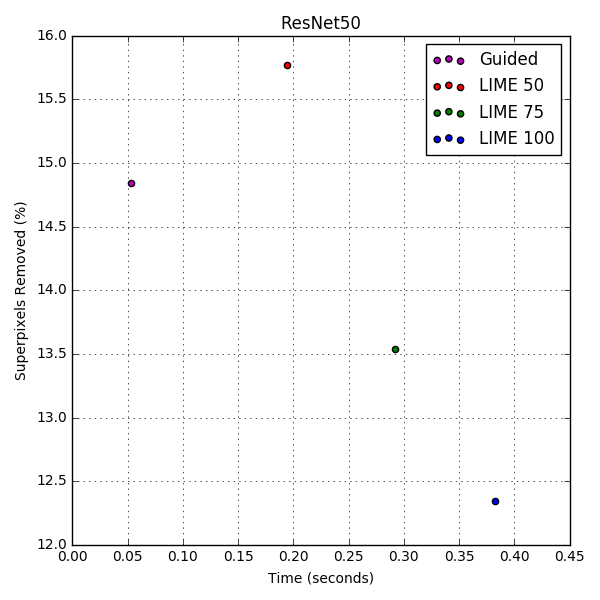}
\includegraphics[width=4cm]{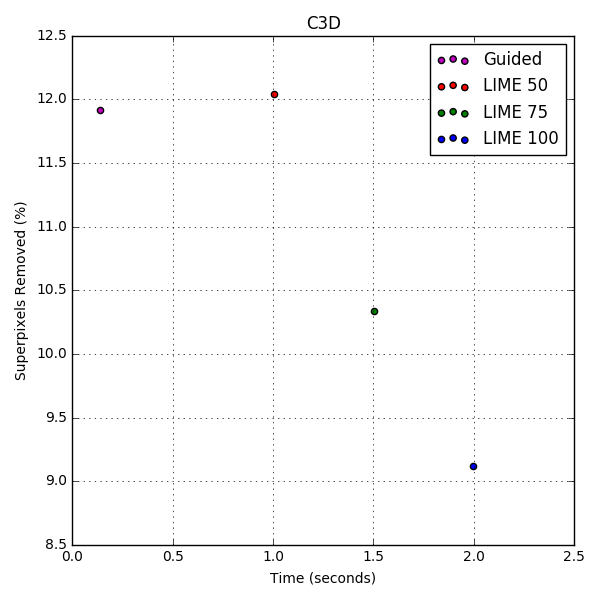}
\caption{Results for the average time taken for weighting segments with guided
backpropagation compared to LIME using 50, 75 and 100 samples.}
\label{figure:timings}
\end{figure}

We position our proposed method as a fast alternative to LIME, therefore we
perform a timing experiment to ascertain how long it takes on average to
generate the superpixel scores for a number of the better performing methods
from our superpixel removal experiments, and for LIME with various number of
samples. For each proposed method we average the time taken to generate the
scores for the images in the validation set of either ImageNet or Kinetics-400
depending on the network. We do not
include the generation of the superpixels within the timings as this is common
to both techniques. For LIME we
use the implementation released by the authors and follow their guidance on
using it with PyTorch. The results for each of the
networks can be found in Figure~\ref{figure:timings}.

From the results in Table~\ref{table:topk_results} it is clear to see that the more samples used with LIME the better the ability to accurately score the superpixels is. However, that comes at the cost of efficiency. Our proposed method is able to approximate LIME between $50$
and $75$ samples with only a single pass through the network, this results in a
much more efficient visualisation time. Of particular note is the C3D network
where our method is able to visualise a $16$ frame temporal volume in $0.13$
seconds compared to $1.0$ seconds and $1.5$ seconds for LIME with 50 and 75
samples respectively.

\section{Conclusion}
In this paper we have introduced a novel method for weighting superpixels using
easy and efficient to obtain pixel values generated with standard visualisation
techniques. We have performed experiments to discover which of these methods
provide the best performance in use and shown how similar results to LIME can be
achieved in a time efficient manner. We have extended the technique to
action recognition networks and shown how it can be applied to offer insight
into networks using a temporal volume as an input.
\section{Future Work}
Going forward there are a number of methods~
\cite{montavon2017explaining, zhang2018top, pmlr-v70-shrikumar17a} for
generating pixel values that would be useful to explore. We would also like to
experiment with varying the number of superpixels generated as the more
superpixels present in an image, the finer grained the explanation becomes, but
the longer it takes for LIME to compute. Finally, expanding the methods use in
action
recognition networks could prove fruitful.

\section{Acknowledgements}
This work is generously funded by BAE Systems and the EPSRC via iCASE award
1852482.

\bibliography{egbib}
\end{document}